\def\year{2020}\relax
\documentclass[letterpaper]{article} 
\usepackage{aaai20}  
\usepackage{times}  
\usepackage{helvet} 
\usepackage{courier}  
\usepackage[hyphens]{url}  
\usepackage{graphicx} 
\usepackage{graphicx}  

\usepackage{amsmath}
\usepackage{amssymb}
\usepackage{booktabs}
\usepackage{algorithm}
\usepackage{algorithmic}
\usepackage{multirow}
\usepackage{bm}
\newcommand{\ra}[1]{\renewcommand{\arraystretch}{#1}}

\title{Age Progression and Regression with Spatial Attention Modules}

\author{
Qi Li\textsuperscript{\rm 1,2}\thanks{denotes equal contribution}\quad~Yunfan Liu\textsuperscript{\rm 1$\ast$}\quad~Zhenan Sun\textsuperscript{\rm 1,2}\footnote{Contact Author} \\
\textsuperscript{\rm 1}Center for Research on Intelligent Perception and Computing, CASIA\\ 
\textsuperscript{\rm 2}National Laboratory of Pattern Recognition, CASIA\\
\{qli, znsun\}@nlpr.ia.ac.cn, yunfan.liu@cripac.ia.ac.cn
}

\begin{document}

\maketitle

\begin{abstract}
Age progression and regression refers to aesthetically rendering a given face image to present effects of face aging and rejuvenation, respectively. Although numerous studies have been conducted in this topic, there are two major problems: 1) multiple models are usually trained to simulate different age mappings, and 2) the photo-realism of generated face images is heavily influenced by the variation of training images in terms of pose, illumination, and background. To address these issues, in this paper, we propose a framework based on conditional Generative Adversarial Networks (cGANs) to achieve age progression and regression simultaneously. Particularly, since face aging and rejuvenation are largely different in terms of image translation patterns, we model these two processes using two separate generators, each dedicated to one age changing process. In addition, we exploit spatial attention mechanisms to limit image modifications to regions closely related to age changes, so that images with high visual fidelity could be synthesized for in-the-wild cases. Experiments on multiple datasets demonstrate the ability of our model in synthesizing lifelike face images at desired ages with personalized features well preserved, and keeping age-irrelevant regions unchanged.
\end{abstract}

\section{Introduction}
The target of age progression and regression is predicting the appearance of a given face at different ages.
An increased research activity in the area of facial aging
simulation is recorded in the last two decades.
Its applications range from social security to digital entertainment, such as cross-age identification and face age editing.
Despite the appealing practical value, the lack of labeled age data of the same subject covering a large time span and the great change in appearance over a long time interval collectively make age progression and regression a difficult problem.

Many approaches have been proposed to tackle this issue.
They can be roughly divided into two types: traditional face aging methods and deep learning based face aging methods.
Traditional face aging methods utilize prototype images or parametric anatomical models to describe
face aging procedure. Burt~et al~\cite{burt1995perception} studied visual cues to age by using facial composites which blend shape and
color information from multiple faces. It is one of the early works for face aging.
O'Toole~et al~\cite{o1997three} suggested that the distinctiveness of a face, which is
defined as its distance from the average
face in a 3D face space, is related to facial age.
Then Tiddeman~et al~\cite{tiddeman2001prototyping} presented
a wavelet based methods for prototyping
facial textures and then transforming it for different age groups.
A craniofacial growth model that characterizes face aging effects is proposed in~\cite{ramanathan2006modeling}.
In the following, they further proposed a two step approach towards face aging in adults~\cite{ramanathan2008modeling}, which comprises
a shape variation model and a texture variation model.

\begin{figure}[t]
\begin{center}
\includegraphics[width=0.95\linewidth]{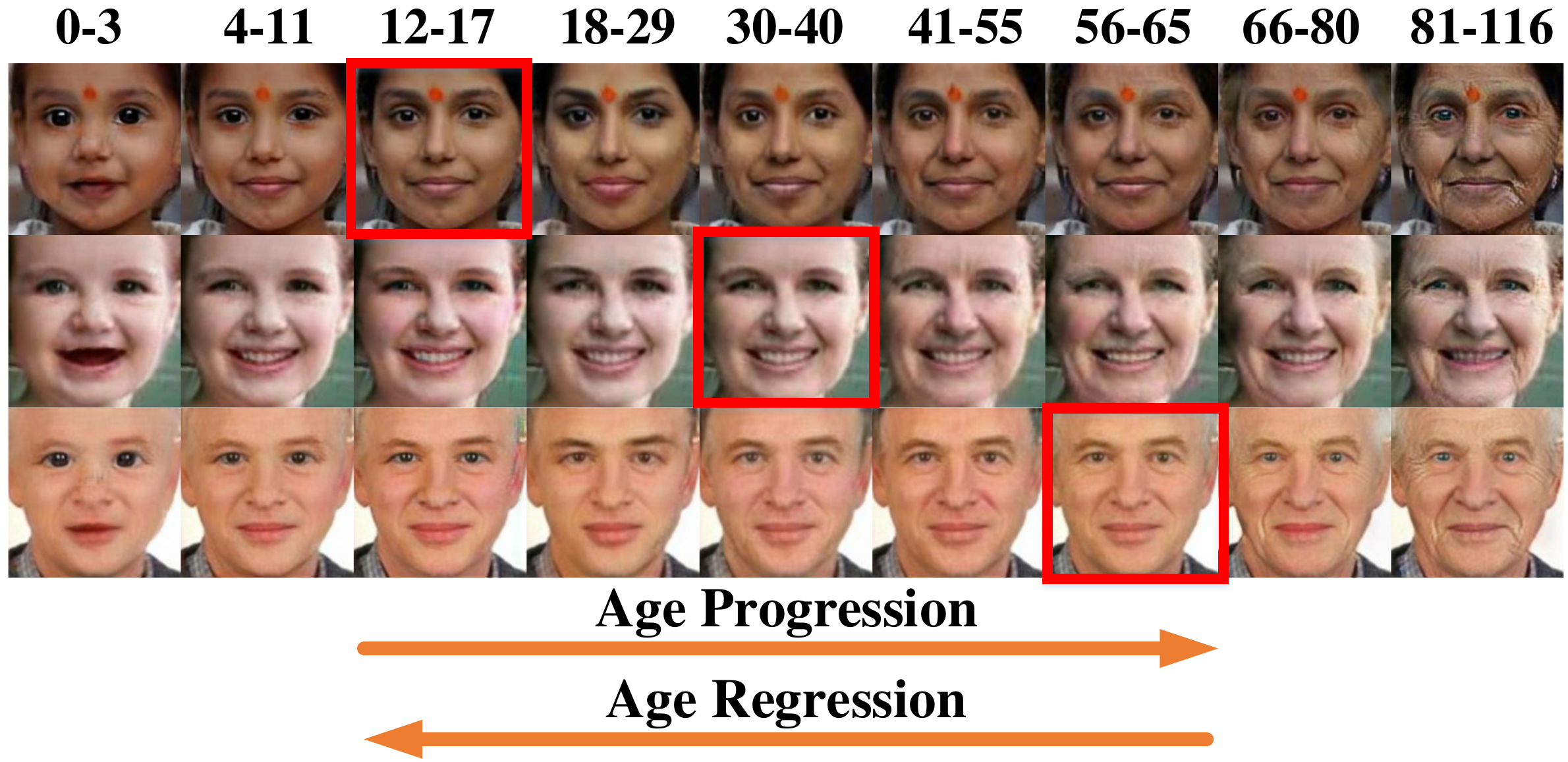}
\end{center}
\caption{Samples results of age progression and regression generated by the proposed method. Red boxes indicate input face images. Clearly, in the age regression process, common patterns of facial appearance change are shared by different subjects (e.g.~bigger eyes and more smooth skin), and this is also true for the age progression process (e.g.~more wrinkles  and deeper laugh lines).}
\label{fig:examples}
\end{figure}

With the success of Generative Adversarial Networks (GANs)~\cite{goodfellow:generative} in generating visually appealing images,
many efforts have been made to solve age progression and regression using GAN-based frameworks.
There have been two key research directions in age progression for GAN-based methods: single model synthesis and multiple model synthesis.
Single model synthesis refers to using only one single conditional image translation model to describe all of the face aging process.
Different age translations are achieved by a single framework with the target age as a prior condition~\cite{zeng2019photo,zhifei:cvpr,antipov:face}.
While multiple model synthesis uses multiple image to image translation models to describe the face aging process.
Deep convolutional GANs are proposed to synthesize face images at given ages~\cite{liu2019attribute,li:global,yang:learning}.
One drawback of single model synthesis methods is that no constraint is enforced to guarantee the target age fulfillment.
As for multiple image synthesis, models have to be trained repeatedly for different source or target ages, which heavily increases the computational cost.


To tackle the above-mentioned issues, in this paper, we propose a conditional GAN based framework to solve age progression and regression simultaneously.
According to examples shown in Figure~\ref{fig:examples}, age progression and regression processes are largely different from each other in terms of image translation patterns.
Therefore, unlike previous works, we propose to model these two processes using two separate generators, each dedicated to one age changing process.
In addition, aging could be considered as adding representative signs (e.g.~wrinkles, eye bags, and laugh lines) to the original input, while rejuvenation is to do the opposite.
That is to say, we would like to limit the modifications to those regions relevant to age changes and ignore the rest for further more accurate processing.
To this end, the spatial attention mechanism  is naturally adopted to constrain image translations, and help to improve the quality of generation results by minimizing the chance of introducing distortions and ghosting artifacts.
In brief, the main contributions of our work could be summarized as follows,
\begin{itemize}
\item We propose to solve age progression and regression in a unified conditional GAN based framework. Particularly, we employ a pair of generators to perform two opposite tasks, face aging and rejuvenation, which take face images and target age conditions as input and synthesize photo-realistic age translated face images.
\item The spatial attention mechanism is introduced to our model to limit modifications to those regions that are relevant to convey the age changes, so that ghosting artifacts could be suppressed and the quality of synthesized images could be improved. To the best of our knowledge, our work is the first to introduce spatial attention mechanism  to face aging and rejuvenation.
\item Extensive experiments on three age databases are conducted to comprehensively evaluate the proposed method. Both qualitative and quantitative results demonstrate the effectiveness of our model in accurately synthesizing face images at desired ages with identity information being well-preserved.
\end{itemize}

\begin{figure}[t]
\begin{center}
\includegraphics[width=85mm,height=88mm]{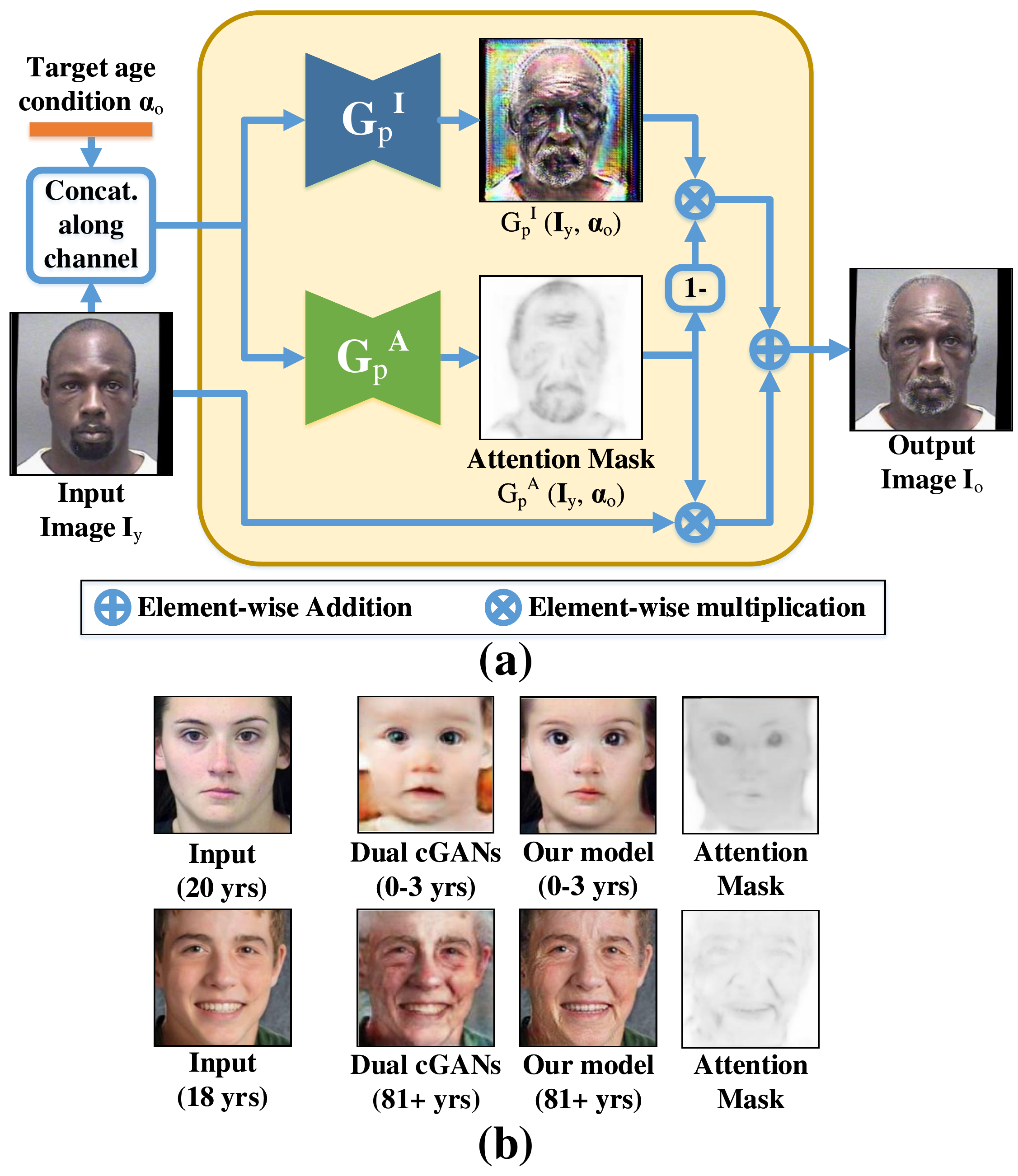}
\end{center}
\caption{The age progressor $G_p$. (a) The detailed structure of $G_p$.
(b) Sample results of our model and Dual cGANs, which demonstrate the effect of attention modules in constraining modifications made to the input.}
\label{fig:generator}
\end{figure}

\section{Related Work}

In the last two decades, many approaches have been proposed to solve age progression and regression, and they could be roughly
divided into three categories: physical model-based methods, prototype-based methods, and deep learning-based methods.

Physical model-based methods simulate the change of facial appearance over time by manipulating the parametric anatomical model of human faces, such as facial muscle and structure~\cite{todd:perception,lanitis:toward,tazoe:facial}.
Thompson~\cite{thompson:growth} suggested that it is possible to use coordinate transformations for altering the shape of biological organisms.
Inspired by this idea, a variant of cardioidal-strain transformations were used to model the growth of human heads in~\cite{todd:perception}.
Todd~\emph{et al}~\cite{todd:perception} modeled the transition of facial appearance
by a variant of cardioidal-strain transformation.
The following works tried to investigate the problem from various biological aspects, such as facial muscles and facial structures~\cite{lanitis:toward,tazoe:facial}. However, physical model-based algorithms are usually very complex and  computational expensive.

As for the prototype-based methods~\cite{suo:compositional,kemelmacher:illumination}, face images are firstly divided into several age groups and an average face is computed as the prototype for each age group.
After that, transition patterns between prototypes are learned and then applied to render effects of age changing.
In~\cite{suo:compositional}, the compositional model represented faces in each age group by a hierarchical and or graph.
Then a Markov process on the graph representation was used to model face aging process.
\cite{kemelmacher:illumination} presented an illumination-aware method for automatic face aging of a single photo by leveraging thousands Internet photos across age groups.
The main problem of prototype-based methods is that the personalized facial texture information are lost when computing the average faces.

In recent years, with the success of GANs in generating high fidelity images, they have been used in many face aging methods to solve the
above problems.
In~\cite{zhifei:cvpr}, a face image is first mapped to a latent vector, and then the latent vector is projected
to the face manifold  conditioned on the age information. Besides, two adversarial
networks are imposed on the encoder and generator in order to generate photo-realistic face images.
To generate more facial details, a GAN-based model with pyramid architecture for face aging was proposed in~\cite{yang:learning}.
Song~et~al~\cite{Song:Dual} integrated the target age condition into the discriminator to supervise the age in the generated images,
and used residual blocks instead of latent vectors at the bottleneck of the generator to preserve image details.
Zeng~et~al formulated face aging as an unsupervised multi-domain image to image translation problem~\cite{zeng2019photo}.

\section{Method}

\subsection{Problem Formulation}

Given a young face image $\mathbf{I}_y$ at age $\bm{\alpha}_y$, we aim to learn an age progressor $G_p$ to realistically translate $\mathbf{I}_y$ into an older face image $\mathbf{I}_o$ at age $\bm{\alpha}_o$ ($\bm{\alpha}_o > \bm{\alpha}_y$), and an age regressor $G_r$ to do the reverse.
To be specific, $G_p$ takes an face image $\mathbf{I}_y$ and the target age condition $\bm{\alpha}_o$ as input, and generates the aged face image $\mathbf{I}_o = G_p(\mathbf{I}_y, \bm{\alpha}_o)$.
However, due to the usage of unpaired aging data, the mapping $\mathbf{I}_y\rightarrow \mathbf{I}_o$ is highly under-constrained and translation patterns other than age progression might be learned.

To deal with this problem, an inverse mapping $\mathbf{I}_o\rightarrow \mathbf{I}_y'$ is usually adopted to reconstruct the input, and the constraint $\mathbf{I}_y'\approx \mathbf{I}_o$ is enforced to regulate the mappings.
It is worth noting that, the inverse mapping is essentially an age regression process, thus is supposed to be naturally accomplished by the age regressor $G_r$, i.e.,~$\mathbf{I}_y' = G_r(\mathbf{I}_o, \bm{\alpha}_y)$.
Similarly, for face rejuvenation, $G_r$ simulates the age regression process and $G_p$ serves as the inverse mapping.
In this way, we integrate $G_p$ and $G_r$ into a single framework, which is a unified solution for both age progression and regression.

\begin{figure*}[ht]\centering
\begin{center}
\includegraphics[width=170mm,height=54mm]{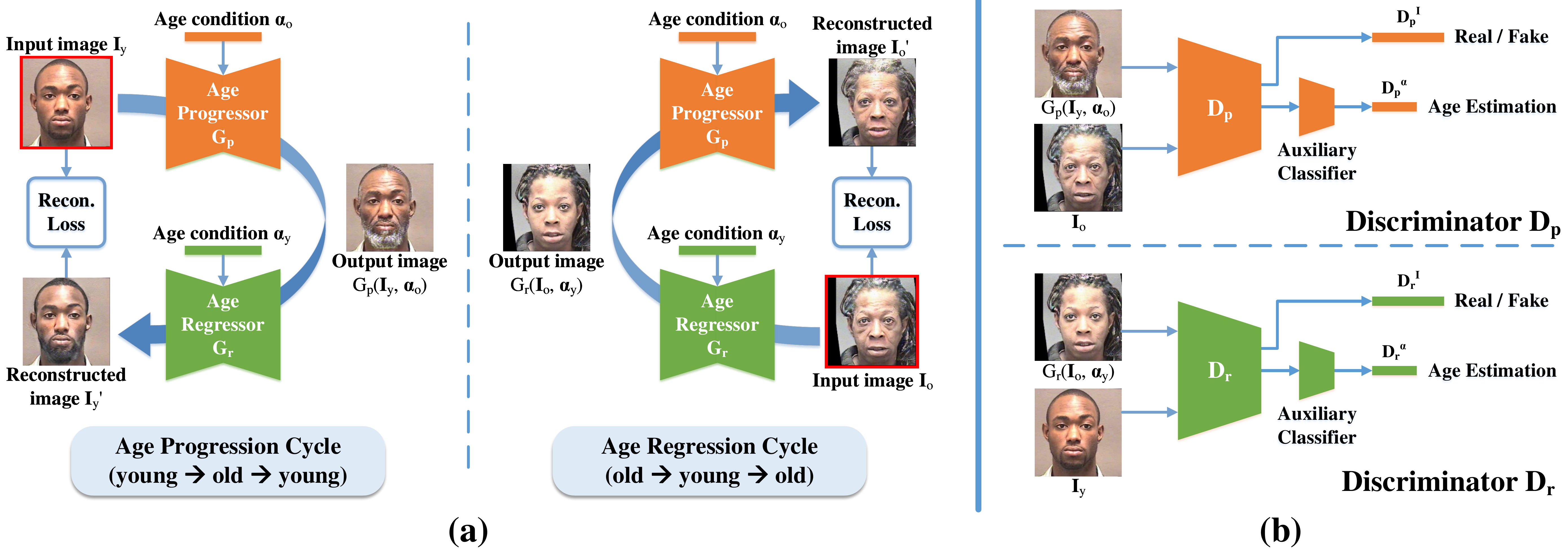}
\end{center}
\caption{The framework of the proposed model. (a): $G_p$ and $G_r$ perform age progression and rejuvenation given the conditional age vector $\bm{\alpha}_o$ and $\bm{\alpha}_y$, respectively. Reconstruction loss is used to ensure that personalized features in the input image is preserved in the output.
(b): $D_p$ and $D_r$ are discriminators aim to distinguish real images from synthetic ones and estimate the age of the input face image, and they are involved in the age progression cycle and regression cycle, respectively.}
\label{fig:overview}
\end{figure*}

The framework of the proposed model is illustrated in Figure~\ref{fig:overview}.
The training process contains two data flow cycles: an age progression cycle and an age regression cycle.
For the age progression cycle, discriminator $D_p$ is employed to encourage the synthesized older face $G_p(\mathbf{I}_y, \bm{\alpha}_o)$ to be as realistic as the real aged face $\mathbf{I}_o$, and the estimated age of $G_p(\mathbf{I}_y, \bm{\alpha}_o)$ to be close to the target age condition $\bm{\alpha}_o$.
Similar for $D_r$ in the age regression cycle.


\subsection{Network Architecture}
In this section, we describe the architecture of the generator and discriminator in detail.
For brevity, in the following discussion, we collectively refer to $G_p$ and $G_r$ as $G$ if there is no need to distinguish the direction, and similar for discriminators $D_p$ and $D_r$ as $D$.

\textbf{Spatial Attention based Generator:} Since the age progressor $G_p$ and age regressor $G_r$ serve equivalent functions, they share the same network architecture.
Therefore, we take $G_p$ for example to describe the detailed architecture, and $G_r$ is different only in terms of input and output.
The structure of $G_p$ is shown in Figure~\ref{fig:generator}.

Most of existing works on face aging use generator with single pathway to predict the whole output image~\cite{zhifei:cvpr,yang:learning,Song:Dual}, where the divergence between the underlying data-generating distribution for the entire image in the source and target age domains are minimized.
Consequently, unintended correspondences between image contents other than age translation (e.g.~background textures) would be inevitably estabilshed, which increases the chance of introducing age-irrelevant changes and ghosting artifacts (as shown in Figure~\ref{fig:generator} (b)).

To solve this problem, we introduce the spatial attention mechanism into our framework, and an additional branch is intergrated to $G_p$ to estimate an attention mask describing the contribution of each pixel to the final output.
To be specific, as shown in Figure~\ref{fig:generator} (a), a fully convolutional network (FCN) $G_p^A$ is used to regress the attention mask, which is fused with the output of another FCN $G_p^I$ to produce the final output.
Mathematically, the image generation process could be described as:
\begin{equation}
\mathbf{I}_{o} = G_p^A(\mathbf{I}_{y}, \bm{\alpha}_o)\cdot \mathbf{I}_{y} + (1 - G_p^A(\mathbf{I}_{y}, \bm{\alpha}_o))\cdot G_p^I(\mathbf{I}_{y}, \bm{\alpha}_o)
\label{eq:attention_module}
\end{equation}
where $\bm{\alpha}_o$ is the one-hot condition vector indicating the target age group, $G_p^A(\mathbf{I}_{y}, \bm{\alpha}_o)\in [0,1]^{H\times W}$ is the attention mask and $G_p^I(\mathbf{I}_{y}, \bm{\alpha}_o)\in \mathbb{R}^{H\times W\times 3}$ models detailed translations within the attended regions.
The greatest advantage of adopting attention mechanism is that the generator could focus only on rendering effects of age changes and irrelevant pixels could be directly retained from the original input, resulting in less distortion and finer image details.

\textbf{Discriminator:} Discriminator $D$ is trained to distinguish synthetic face images from real ones and check whether a generated face image belongs to the desired age group.
The architecture of $D$ we used is similar to PatchGAN~\cite{isola:pix2pix} which has achieved success in a number of image translation tasks.
Concretely, we use a series of six convolutional layers with increasing number of $4\times 4$ filters, and each layer is followed by a Leaky ReLU unit.
In addition, to check whether a synthetic image belongs to the age group represented by the corresponding target age condition, we append an auxiliary fully connected network to the top of $D$ to predict the age of the face image.
Given an input image $\mathbf{I}$, we denote the output of the convolutional layers by $D^I(\mathbf{I})$ and the result of age estimation by $D^\alpha(\mathbf{I})$.

\begin{figure*}[t]
\centering
\includegraphics[width=1.0\linewidth]{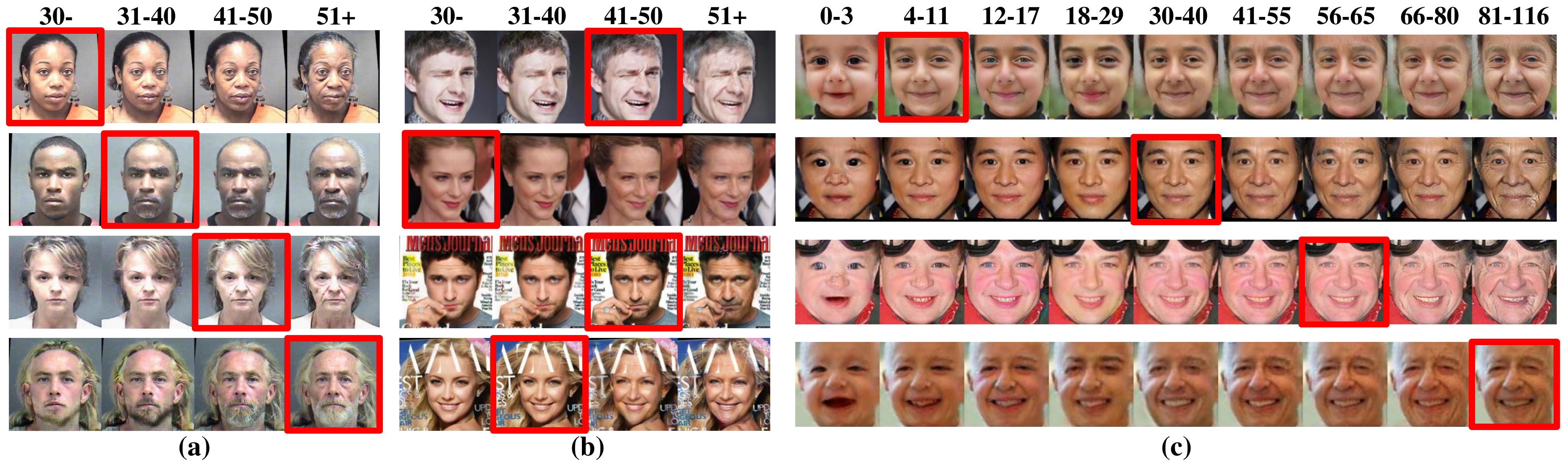}
\caption{Sample results generated by the proposed model. (a) Results on Morph (first two rows) and CACD (last two rows); (b) Results on UTKFace. Red box indicates the input face image for each age progression/rejuvenation sequence. Zoom in for better details.}
\label{fig:vis_results}
\end{figure*}

\begin{figure*}[t]
\centering %
\includegraphics[width=1.0\linewidth]{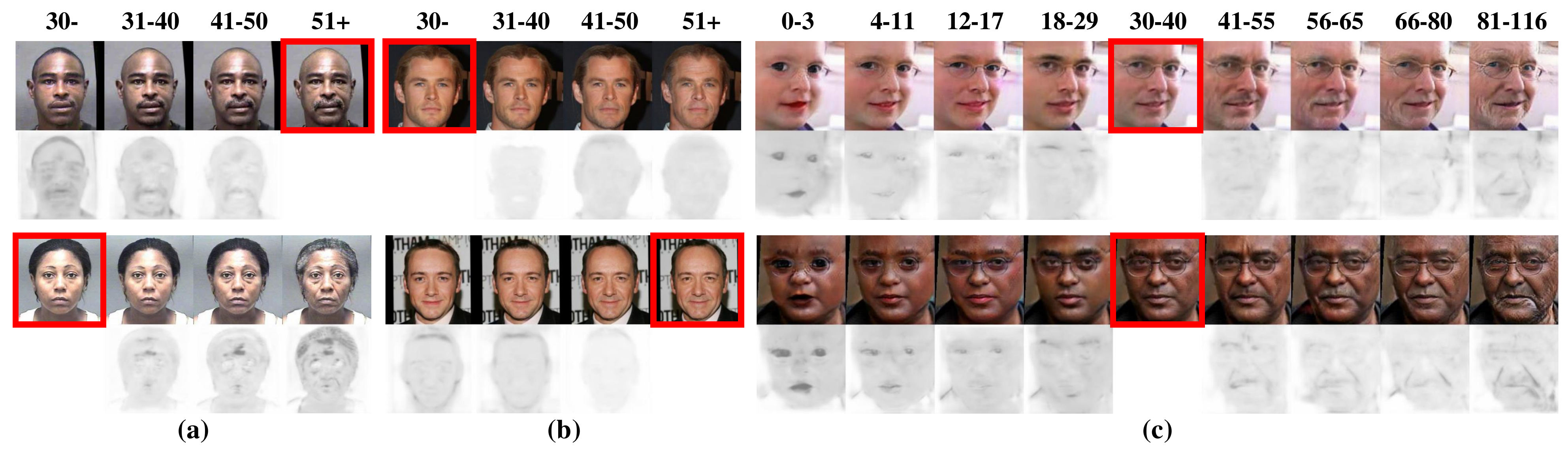}
\caption{Illustration of generation results (first row) and the corresponding attention maps (second row) for each age progression/regression sequence on (a) Morph, (b) CACD, and (c) UTKFace. Red boxes indicate input face images. For the attention maps, darker regions suggest those areas of the associated face image receive more attention in the generation process, and brighter regions indicate that more information are retained from the original input image. Zoom in for better details.}
\label{fig:vis_attention_map}
\end{figure*}


\subsection{Loss Function}
The loss function of the proposed model contains four parts: an adversarial loss to encourage the distribution of generated images to be indistinguishable from that of real images, a reconstruction loss to preserve personalized features, an attention activation loss to prevent saturation, and an age regression loss to measure the target age fulfillment.

\textbf{Adversarial Loss:} Adversarial loss describes the objective of a minimax two-player game between the generator $G$ and the discriminator $D$, where $D$ aims to classify real images from fake ones and $G$ attempts to fool $D$ with lifelike synthetic images.
Unlike regular GANs, least square adversarial loss is employed in our model to improve the quality of generated images and stabilize the training process. Mathematically, the objective of adversarial loss could be formulated as follows,
\begin{align}
\mathcal{L}_{GAN} = &\: \mathbb{E}_{\mathbf{I}_y}[(D^I_p(G_p(\mathbf{I}_y, \bm{\alpha}_o)) - 1)^2] \nonumber\\
                  + &\: \mathbb{E}_{\mathbf{I}_o}[(D^I_p(\mathbf{I}_o) - 1)^2] + \mathbb{E}_{\mathbf{I}_y}[D^I_p(G_p(\mathbf{I}_y, \bm{\alpha}_o))^2] \nonumber\\
                  + &\: \mathbb{E}_{\mathbf{I}_o}[(D^I_r(G_r(\mathbf{I}_o, \bm{\alpha}_y)) - 1)^2] \nonumber\\
                  + &\: \mathbb{E}_{\mathbf{I}_y}[(D^I_r(\mathbf{I}_y) - 1)^2] + \mathbb{E}_{\mathbf{I}_o}[D^I_r(G_r(\mathbf{I}_o, \bm{\alpha}_y))^2]
\label{eq:adversarial_loss}
\end{align}

\textbf{Reconstruction Loss:} With the adversarial loss, $G$ learns to generate lifelike face images at the target age.
However, these is no guarantee that personalized features in the input image are preserved in the output since no ground-truth supervision is available.
Therefore, a reconstruction loss is employed to penalize the difference between the input image and its reconstruction, which could be formulated as
\begin{align}
\mathcal{L}_{recon} = &\: \mathbb{E}_{\mathbf{I}_y}[\: \|G_r(G_p(\mathbf{I}_y, \bm{\alpha}_o)) - \mathbf{I}_y\|_1 \:] \nonumber \\
                    + &\: \mathbb{E}_{\mathbf{I}_o}[\: \|G_p(G_r(\mathbf{I}_o, \bm{\alpha}_y)) - \mathbf{I}_o\|_1 \:]
\label{eq:reconstruction_loss}
\end{align}
Here we use the L1-norm to encourge less blurred results.

\textbf{Attention Activation Loss:} In Equation~(\ref{eq:adversarial_loss}) and~(\ref{eq:reconstruction_loss}), simply setting $G_p$ and $G_r$ to identity mapping would minimize these loss terms, which is definitely not what we expected.
In this case, as shown in Equation~(\ref{eq:attention_module}), all elements in $G^A(\mathbf{I}_{in}, \bm{\alpha})$ saturate to 1 and thus $\mathbf{I}_{out}=\mathbf{I}_{in}$.
To prevent this, an attention activation loss is used to constraint the total activation of the attention mask, which could be written as
\begin{equation}
\mathcal{L}_{actv} = \mathbb{E}_{\mathbf{I}_y}[\: \|G_p^A(\mathbf{I}_y, \bm{\alpha}_o)\|_2 \:] + \mathbb{E}_{\mathbf{I}_o}[\: \|G_r^A(\mathbf{I}_o, \bm{\alpha}_y)\|_2 \:]
\end{equation}

\textbf{Age Regression Loss:} Apart from being photo-realistic, synthesized face images are also expected to satisfy the target age condition.
Therefore, an age regression loss is adopted to force generators to reduce the error between estimated ages and target ages, which could be expressed as
\begin{align}
\mathcal{L}_{reg} = &\: \mathbb{E}_{\mathbf{I}_y}[\: \|D_p^\alpha(G_p(\mathbf{I}_y, \bm{\alpha}_o)) - \bm{\alpha}_o\|_2 \:] \nonumber \\
                  + &\: \mathbb{E}_{\mathbf{I}_y}[\: \|D_p^\alpha(\mathbf{I}_y) - \bm{\alpha}_y\|_2 \:] \nonumber \\
                  + &\: \mathbb{E}_{\mathbf{I}_o}[\: \|D_r^\alpha(G_r(\mathbf{I}_o, \bm{\alpha}_y)) - \bm{\alpha}_y\|_2 \:] \nonumber \\
                  + &\: \mathbb{E}_{\mathbf{I}_o}[\: \|D_r^\alpha(\mathbf{I}_o) - \bm{\alpha}_o\|_2 \:]
\label{eq:age_regression_loss}
\end{align}
By optimizing Equation~(\ref{eq:age_regression_loss}), the auxiliary regression network $D^\alpha$ gains the age estimation ability, and the generator $G$ is encouraged to accurately render fake faces at the desired age.

\textbf{Overall Loss:} The final full loss function of the proposed model could be formulated as the linear combination of all previously defined losses:
\begin{equation}
\mathcal{L} = \mathcal{L}_{GAN} + \lambda_{recon}\mathcal{L}_{recon} + \lambda_{actv}\mathcal{L}_{actv} + \lambda_{reg}\mathcal{L}_{reg}
\end{equation}
where $\lambda_{recon}$, $\lambda_{actv}$, and $\lambda_{reg}$ are coefficients balancing the relative importance of each loss term.
Finally, $G_p$, $G_r$, $D_p$, and $D_r$ are solved by optimizing:
\begin{equation}
\min_{G_p, G_r} \max_{D_p, D_r} \mathcal{L}
\end{equation}

\begin{figure*}[t]
\centering
\includegraphics[width=1.0\linewidth]{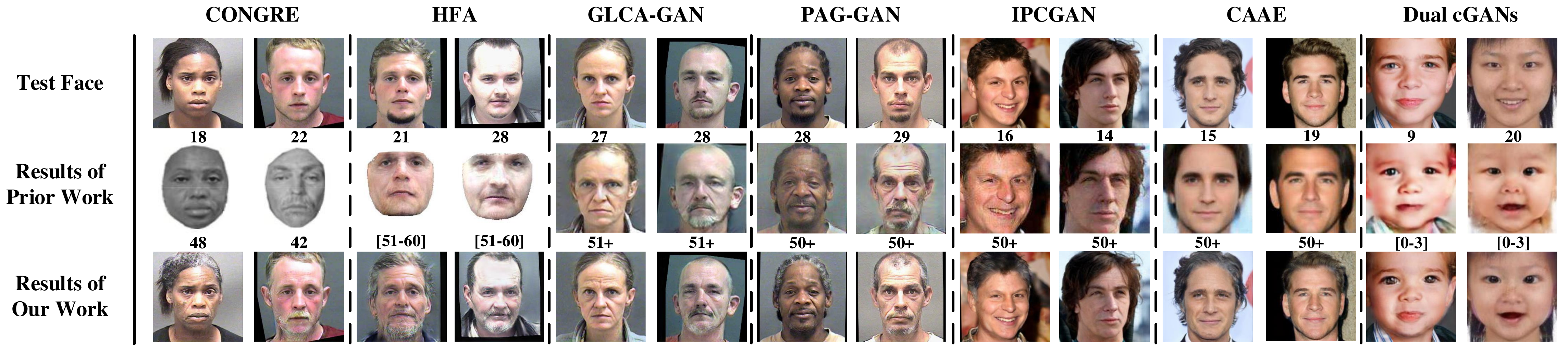}
\caption{Performance comparison with prior work on Morph (zoom in for a better view of image details). The second row shows the results of prior work, where seven methods are considered and two sample results are presented for each. Results generated by our method are shown in the last row. Ages as labeled underneath, and results of our work and the associated prior work are in the same age group.}
\label{fig:vis_compare}
\end{figure*}

\subsection{Relation To Previous Work}
In this part, we emphasize the differences between our method and several state-of-the-art multiple model synthesis methods: GANimation~\cite{pumarola2018ganimation}, CAAE~\cite{zhifei:cvpr}, IPCGAN~\cite{wang:face_aging} and Dual cGANs~\cite{Song:Dual}.
First of all, we propose to model the age progression and regression process separately using a pair of dedicated generators and train them jointly in a cyclic manner. However, GANimation, CAAE and IPCGAN adopt non-cyclic models to do image translations directly. Besides, the spatial attention mechanism is proposed into our method  to aesthetically and meticulously render the given face image to present the effects of age changing. The detailed architecture of our generator and the way conditional information is integrated are also different with GANimation, CAAE, IPCGAN and Dual cGANs. Moreover, unlike most other methods using simple age vector concatenation (Dual cGANs) or large pre-trained age classifier (IPCGAN), an auxiliary lightweight age regressor is employed in the discriminator and to ensure target age fulfillment more efficiently.

\section{Experiments}

\subsection{Datasets}
Three publicly available face aging datasets, Morph~\cite{ricanek:morph}, CACD~\cite{chen:face}, and UTKFace~\cite{zhifei:cvpr} are used in our experiments.
Following~\cite{yang:learning} and~\cite{li:global}, we divide images in Morph and CACD into 4 age groups (30-, 31-40, 41-50, 51+), and UTKFace into 9 age groups (0-3, 4-11, 12-17, 18-29, 30-40, 41-55, 56-65, 66-80, and 81-116) according to~\cite{Song:Dual}.
For each dataset, we randomly select 80\% images for training and the rest for testing, and ensure that these two sets do not share images of the same subject.

\subsection{Implementation Details}
All face images are aligned according to both eyes and then resized to $256\times 256$.
We train our model for 30 epochs with batchsize of 24, using the Adam optimizer with learning rate set to 1e-4.
Optimization over generators is performed every 5 iterations of discriminators.
As for the balancing hyper-parameters $\lambda_{recon}$, $\lambda_{actv}$, and $\lambda_{reg}$, we first initialize them to make all losses to be of the same order of magnitude as the adversarial loss $\mathcal{L}_{GAN}$, then divide them by 10 except for $\lambda_{reg}$ to emphasize the importance of accurate age simulation.

\subsection{Qualitative Results}
Sample results of age progression and regression are shown in Figure~\ref{fig:vis_results}.
Although input faces cover a wide range of population in terms of age, race, gender, pose and expression, the model successfully renders photo-realistic and diverse age changing effects.
In addition, it could be observed that identity permanence is well-achieved in all generated face images.

Figure~\ref{fig:vis_attention_map} displays the attention masks for sample generation results.
Note that the network has learned to focus its attention onto face regions most relevant to representative signs of age changes (e.g., wrinkles, laugh
lines, mustache) in an unsupervised manner, and leave other parts of the image unattended.
Figure~\ref{fig:vis_attention_map} (b) shows how attention maps help to deal with occlusions and complex background, that is, by assigning lower attention scores to pixels in these regions.
This allows pixels in unattended areas to be directly copied from the original input image, which improves the visual quality of generated face images, especially for in-the-wild cases.
The proposed method is highly scalable as it could be naturally extended to different age span and age group divisions, as shown in Figure~\ref{fig:vis_attention_map} (c).

To demonstrate the effectiveness of our model, we compare the proposed method with several benchmark approaches: CONGRE~\cite{suo:concatenational}, HFA~\cite{yang:face}, GLCA-GAN~\cite{li:global}, Pyramid-architectured GAN (referred to as PAG-GAN)~\cite{yang:learning}, IPCGAN~\cite{wang:face_aging}, CAAE~\cite{zhifei:cvpr}, and Dual cGANs~\cite{Song:Dual}.
Note that PAG-GAN and GLCA-GAN are multiple model synthesis methods. They need to be trained repeatedly for each age mapping, and the computational cost of these methods is $N \times N$ times as much as ours, where $N$ is the number of age groups.
The comparison results are shown in Figure~\ref{fig:vis_compare}.
It is clear that traditional face aging methods, CONGRE and HFA, only render subtle aging effects within tight facial area, while our method could simulate the aging process on the entire face image.
As for GAN-based methods, GLCA-GAN, PAG-GAN, and IPCGAN, our model is better at suppressing ghosting artifacts and color distortion as well as rendering enhanced aging details.
This is because the attention module enables the model to retain pixels in areas irrelevant to age changes instead of re-estimating them, which avoids introducing additional noise and distortions.
This is also confirmed by the comparison between our method and Dual cGANs, as background, hair regions, and face boundaries are better maintained in the results of our model.

\begin{table*}[ht]\centering
\ra{1.1}
\begin{tabular}{@{}l rr c rr c rr@{}}
\toprule
                  &\multicolumn{2}{c}{Morph}               &\phantom{a}     &\multicolumn{2}{c}{CACD}     &\phantom{a}             &\multicolumn{2}{c}{UTKFace}  \\
\cmidrule{2-3}\cmidrule{5-6}\cmidrule{8-9}
                  & Age Est. Error & Veri. Rate (\%) &                      & Age Est. Error     & Veri. Rate (\%) &               & Age Est. Error      & Veri. Rate (\%) \\
\midrule
CAAE              & $10.34\pm5.63$ & $34.83\, (71.75)$  &                   & $5.16\pm7.08$      & $3.59\, (59.90)$   &            & $11.64\pm10.41$     & $8.07\, (59.40)$ \\
IPCGAN            & $1.74\pm7.44$  & $99.86\, (94.04)$  &                   & $8.11\pm9.69$      & $99.19\, (91.60)$  &            & $7.51\pm11.66$      & $97.32\, (92.63)$ \\
Dual cGANs        & $2.44\pm6.03$  & $99.99\, (93.15)$  &                   & $3.28\pm8.01$      & $99.88\, (93.85)$  &            & $7.01\pm11.29$      & $92.29\, (87.73)$ \\
w/o OI            & $1.58\pm6.50$  & $\mathbf{100.00}\, (95.48)$ &          & $1.92\pm9.36$      & $99.91\, (96.30)$  &            & $5.49\pm11.48$      & $97.08\, (92.38)$ \\
w/o ATT           & $\mathbf{1.53}\pm6.52$  & $\mathbf{100.00}\, (95.50)$ & & $1.89\pm9.07$      & $\mathbf{99.92}\, (96.52)$ &    & $6.81\pm12.29$      & $97.24\, (92.47)$ \\
Ours              & $\mathbf{1.53}\pm6.50$  & $\mathbf{100.00}\, (95.67)$ & & $\mathbf{1.78}\pm7.53$  & $\mathbf{99.92}\, (96.13)$ &  & $\mathbf{4.77}\pm10.59$  & $\mathbf{98.10}\, (92.74)$ \\
\bottomrule
\end{tabular}
\caption{Comparison of quantitative measurements, age estimation error (Age Est. Error) and face verification rate (Veri. Rate) on three face age datasets. Due to the limited space, we only report the mean value and standard deviation of age estimation error computed on all age groups. Verification scores are shown in brackets to provide more information for comparison between different methods.}
\label{tab:est_age}
\end{table*}

\subsection{Quantitative Evaluations}
In this subsection, we report quantitative evaluation results on age translation accuracy and identity preservation.
For age translation accuracy, we calculate the error between estimated ages of real and fake face images, and for identity preservation, face verification rates are reported along with verification scores.
The threshold is set to 76.5@FAR=1e-5 for all identity preservation experiments.
For the sake of fairness, we compare our method with state-of-the-art approaches CAAE, IPACGAN, and Dual cGANs, which all attempt to solve age progression and regression via a single unified framework.
To be objective, all metrics are estimated by the publicly available online face analysis tools of Face++~\footnote{Face++ Research Toolkit (http://www.faceplusplus.com).}, so that results are more objective and reproducible compared to those obtained by user study.

According to results shown in Table~\ref{tab:est_age}, our method achieves the best performance in both age translation accuracy and identity preservation on all three datasets, and outperforms other methods by a clear margin.
CAAE produces over-smoothed face images with subtle changes, leading to large errors in estimated ages and low face verification scores.
For IPCGAN, our method achieves much higher age translation accuracy than IPCGAN does on CACD and UTKFace.
This is because IPCGAN re-estimates all pixels in the output images, thus the chance of introducing distortions is increased, especially for in-the-wild images in CACD and UTKFace.
In addition, compared to IPCGAN, our method generates images with higher resolution ($256\times 256$ vs. $128\times 128$) and do not require pre-trained age classifier and identity feature extractor, which significantly simplifies the training process.
The major difference between our method and Dual cGANs is the adoption of spatial attention modules, and this explains why our method outperforms Dual cGANs under both metrics, which again demonstrates the effectiveness of attention mechanism in improving the quality of generated images.

\subsection{Ablation Study}
Experiments are conducted to fully explore the contributions of adopting attention modules (ATT) and using ordered input (OI) in simulating age translations.
According to results shown in Table~\ref{tab:est_age}, either removing attention modules or using unordered training pairs would cause performance drops for both metrics, which clearly demonstrate that both attention modules and ordered input are critical to reach the optimal performance.

In addition, it could be observed that the advantage brought by ATT and OI increases as it goes from controlled training samples (Morph) to in-the-wild face images (CACD and UTKFace), and from relatively concentrated face attributes (white and black people of 20 to 60 years old in Morph) to diverse data distributions (face images of up to 5 races covering an larger age span in CACD and UTKFace).
This is mainly because that attention modules save the network from being interfered by variations irrelevant to age changes, which brings more advantages to in-the-wild cases than controlled ones.
In addition, arranging input training samples by age enables the two generators to focus only on one direction of age change, which facilitates the convergence of the overall model and improves the quality of generation results.

\subsection{Generalization Ability Study}

\begin{figure}[t]
\begin{center}
\includegraphics[width=1.0\linewidth]{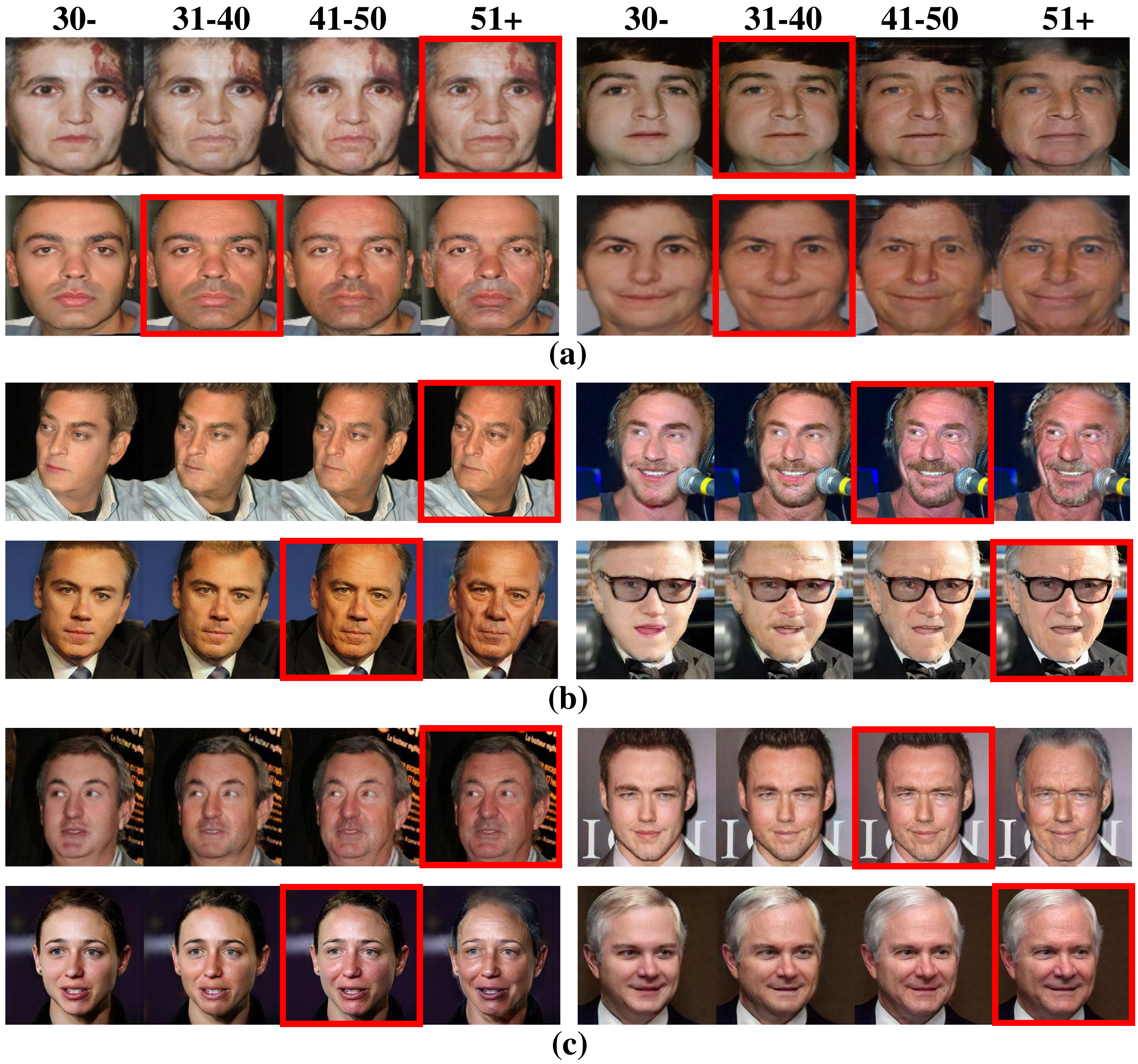}
\end{center}
\caption{Sample results on (a) FG-NET, (b) CelebA, and (c) IMDB-WIKI dataset generated by the model trained on CACD. Red boxes indicate input images.}
\label{fig:generalization}
\end{figure}

We evaluate the generalization ability of our method by applying the model trained on CACD to images from FG-NET~\cite{lanitis:toward}, CelebA~\cite{liu:faceattributes}, and IMDB-WIKI~\cite{Rothe:DEX}.
For those images without age labels, apparent ages are detected using the Face++ API.
Sample results shown in Figure~\ref{fig:generalization} demonstrate that our model generalizes well to face images with different data distributions, and accurate age labels are not strictly required.
Note how the occlusions (e.g.~scars, glasses and microphones) and background in the input images are preserved in the output.

\section{Conclusion}
This paper proposes a conditional GAN-based model to solve age progression and regression simultaneously.
Based on the patterns of facial appearance change in the age progression and regression process, we propose to use a pair of generators to simulate these opposite processes: face aging and rejuvenation.
In addition, the spatial attention mechanism is introduced in our work to present the effects of age changing.
As a result, our model learns to focus only on those regions of the image relevant to age translations, making it robust to distracting environmental factors, such background with complex textures.
Extensive experimental results demonstrate the effectiveness of our method in achieving accurate age translation and successful identity preservation, especially for in-the-wild cases.

\bibliographystyle{aaai}
\bibliography{egbib}

\end{document}